\documentclass[runningheads]{llncs}
\usepackage[T1]{fontenc}
\usepackage{appendix}
\usepackage{graphicx}
\usepackage{bbding}
\usepackage{float}
\usepackage{latexsym}
\usepackage{array}
\usepackage{booktabs} 
\usepackage{multirow}

\usepackage{microtype}

\usepackage{inconsolata}
\usepackage{amsfonts,amssymb}
\usepackage{amsmath}
\begin{document}
\title{Enhanced Multimodal Aspect-Based Sentiment Analysis by LLM-Generated Rationales}
%
\author{Jun Cao \inst{1,2}\orcidID{0009-0006-0344-520X} \and
Jiyi Li\inst{2}\orcidID{0000-0003-4997-3850} \and
Ziwei Yang\inst{1,2}
\and
Renjie Zhou\inst{1 (}\Envelope\inst{)}}
\authorrunning{J.Cao et al.}
%
\institute{
School of Computer Science and Technology, Hangzhou Dianzi University,
Hangzhou 310018, China
\\ \email{\{221050372, 221050378, rjzhou\}@hdu.edu.cn}\and Faculty of Engineering, Graduate Faculty of Interdisciplinary Research,
University of Yamanashi, Kofu 400-8511, Japan \\ \email{jyli@yamanashi.ac.jp} 
}
\maketitle            
\begin{abstract}
 There has been growing interest in Multimodal Aspect-Based Sentiment Analysis (MABSA) in recent years. Existing methods predominantly rely on pre-trained small language models (SLMs) to collect information related to aspects and sentiments from both image and text, with an aim to align these two modalities. However, small SLMs possess limited capacity and knowledge, often resulting in inaccurate identification of meaning, aspects, sentiments, and their interconnections in textual and visual data. On the other hand, Large language models (LLMs) have shown exceptional capabilities in various tasks by effectively exploring fine-grained information in multimodal data. However, some studies indicate that LLMs still fall short compared to fine-tuned small models in the field of ABSA. Based on these findings, we propose a novel framework, termed LRSA, which combines the decision-making capabilities of SLMs with additional information provided by LLMs for MABSA. Specifically, we inject explanations generated by LLMs as rationales into SLMs and employ a dual cross-attention mechanism for enhancing feature interaction and fusion, thereby augmenting the SLMs' ability to identify aspects and sentiments. We evaluated our method using two baseline models, numerous experiments highlight the superiority of our approach on three widely-used benchmarks, indicating its generalizability and applicability to most pre-trained models for MABSA.

\end{abstract}
\section{Introduction}
There has been growing interest in Multimodal Aspect-Based Sentiment Analysis (MABSA) as a complex fine-grained task within sentiment analysis \cite{lv2021aspect,2}. The MABSA task encompasses three primary objectives: (1) Multimodal Aspect Term Extraction (MATE), which is focused on identifying aspect terms with sentiment polarity from a sentence within an image-text pair; (2) Multimodal Aspect-Based Sentiment Classification (MASC), which focuses on determine the sentiment polarity for each identified aspect; (3) Multimodal Aspect-Based Sentiment Analysis (MABSA), which entails jointly extracting all aspect terms and predicting their sentiment polarity from the image-text pair \cite{2}.

Despite significant advancements, aspect extraction and sentiment polarity prediction in this fine-grained sentiment analysis scenario with multimodal information remain challenging for current models. Firstly, the semantic complexity of sentences poses difficulties in identifying aspects and comprehending their corresponding sentiment. Secondly, images often contain abundant detailed information, and models frequently struggle to focus on all key information in the image or mistakenly pay attention to irrelevant areas, introducing noise. Lastly, it is challenging for the model to capture the overall connection between images and text, particularly when judging the connection between more specific image regions and textual vocabulary.
Recent approaches often employ pre-trained small language models (SLMs) to provide representations and understanding of image-text pairs \cite{3,4}. These approaches also incorporate various modules to assist the small models in comprehending information from image-text pairs. Other enhancement methods include contrastive learning and image-text matching to align semantics and mitigate modality gaps \cite{yang2022cross}. While SLMs have demonstrated improvements, their knowledge and limitations also influence the further improvement of the models. Incorporating external knowledge is essential to strengthen the SLMs' understanding of image-text pairs. Some methods utilize image captions generated by ClipCap \cite{5,mokady2021clipcap} as image prompts to assist image comprehension. However, this external information is often simplistic and crude, as image captions usually lack a relationship with sentiments and provide limited assistance in fine-grained aspect extraction and sentiment judgment.
\begin{figure}\vspace{-1em}
    \centering
    \includegraphics[width=1\linewidth]{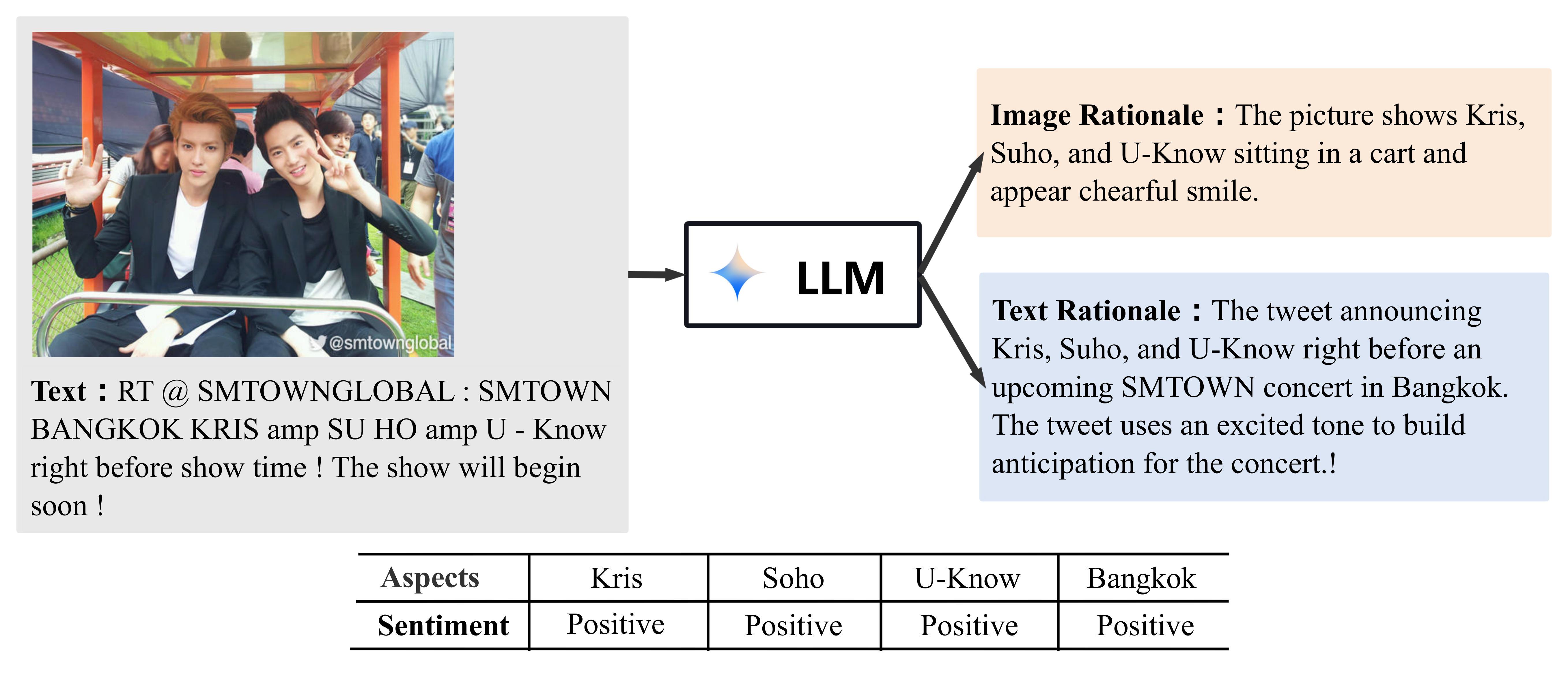}
    \caption{An example of the LLM processing image-text pairs to generate the explanation of images and text.}
    \label{fig1}
\end{figure}
Large Language Models (LLMs) possess powerful abilities to understand, track, and generate complex language \cite{openai2022chatgpt,anthropic2023claude}. They have made significant impressions in various tasks \cite{wei2022emergent} and are considered promising solutions for general-purpose tasks \cite{ma2023large}. LLMs are typically trained on large-scale text corpora, granting them richer knowledge and the ability to mine fine-grained information, subtle patterns, and relationships from data. Nevertheless, studies have demonstrated that LLMs' performance on the ABSA task is not as favorable as fine-tuned small models. We believe that LLMs are yet incapable of selecting and integrating received information to draw conclusions and make judgments in this fine-grained task. However, by employing SLMs' flexible task-specific learning and utilizing LLMs' generated information as informative rationales, we can combine the strengths of small and large models. LLM leverages its robust knowledge repository, superior information retrieval, as well as reasoning capabilities to generate insightful rationales. These rationales serve as external sources that the small model can refer to for making informed judgments. With these considerations, we propose a novel framework LRSA that combines the decision-making abilities of SLMs with the additional information provided by LLMs for MABSA. Specifically, we use LLMs to generate rationales for understanding image-text pairs and their connection as shown in Figure \ref{fig1}, and to inject this information in SLMs as the connection of the SLMs and LLMs. Inside SLMs, to further improve the model's performance, we employ a dual cross-attention mechanism to fuse the rationales with the image-text pair, emphasizing the useful information related to the current task and eliminating unnecessary noise. The image and text features refer to the rationale features in the fusion process to enhance them, making the fused feature representation more comprehensive and accurate, thus improving the understanding ability and predictive performance of the SLM. In the result generation stage, we connect the fused feature of the image-text pairs with the rationale feature to provide more information for the SLMs to make judgments.
Our contributions can be summarized as follows: 

1) We propose a novel framework LRSA that combines the decision-making capabilities of SLMs with the additional information provided by LLMs for MABSA, by using rationales generated by LLMs for SLMs to reference. We deeply integrate LLMs in the field of MABSA. 

2) To better capture fine-grained information and enhance feature representation, we employ a dual cross-attention mechanism to separately fuse image features with the rationales features of LLMs' understanding of images, and text features with the rationales features of LLMs' understanding of texts. 

3) We evaluated our method using two baseline models, experimental results on two benchmark datasets show that our method surpasses those of the baseline models, indicating its generalizability and applicability to most pre-trained models for MABSA.
\section{Related works}
\subsection{Aspect-based Sentiment Analysis}
Aspect-based Sentiment Analysis (ABSA) aims to predict the sentiment polarity of aspect terms within sentences as a fine-grained task. Benefiting from the excellent performance of language models like BERT \cite{devlin2018bert} and RoBERTa \cite{liu2019roberta} in NLP, most recent works use pre-trained language models to to capture the semantic relationship between a given aspect and its context \cite{rietzler2019adapt,liang2022aspect}.
In addition, constructing aspect-oriented dependency trees can assist in linking aspect words with opinion words and learning the syntactic feature representation of aspects to to enhance the performance of language models further\cite{wang2020relational,tian2021aspect}.
More recently, studies have also adopted end-to-end models for extracting all aspects and sentiment elements in triplets.-
\cite{peng2020knowing,chen2022enhanced}
\subsection{Multimodal Aspect-based Sentiment Analysis}
As social media rapidly expands, Multimodal Aspect-Based Sentiment Analysis has garnered significant research interest in recent years. Ju et al. \cite{2} were the first to implement MABSA within a cohesive framework, handling two subtasks in an end-to-end approach to simultaneously extract aspect terms along with associated emotions. They also developed an auxiliary tool for detecting cross-modal relationships to effectively control the use of visual information. Ling et al. \cite{3} developed a generative multimodal framework based on BART to facilitate cross-modal alignment which combines vision-language pre-training with subsequent tasks. Yang et al. \cite{yang2022cross} introduced a cross-modal multi-task transformer (CMMT) for MABSA, which dynamically controls the contribution of visual information to different aspects by taking into account the contribution of images when the confidence of the pure text prediction result is lower. Zhou et al. \cite{4} introduced an aspect-oriented approach aimed at identifying both aspect-related semantic and sentiment information to mitigate visual and textual noise in complex image-text interactions. Yang et al. \cite{5} explored the MABSA task in the case of a few samples and presented an innovative generative multimodal cue model tailored for MABSA, executing three MABSA-related tasks using a relatively modest dataset of labeled multimodal samples.The above studies mainly concentrate on effectively obtaining representations for both text and images while also capturing the interactive fusion between these two modalities. However, pre-training models alone are not enough. The existing traditional pre-training models have limited capabilities and knowledge and are often unable to precisely identify the meaning, aspects, and sentiment within both text and images. The understanding of the connection between text and image is also lacking. It needs to be strengthened through auxiliary modules. To tackle the aforementioned issues, we propose an innovative framework LRSA that combines the decision-making capabilities of SLMs with the additional information provided by LLMs for MABSA, using the image-text rationale generated by LLM to improve the understanding and prediction performance of SLM.
\begin{figure*}
    \centering
    \includegraphics[width=1\linewidth]{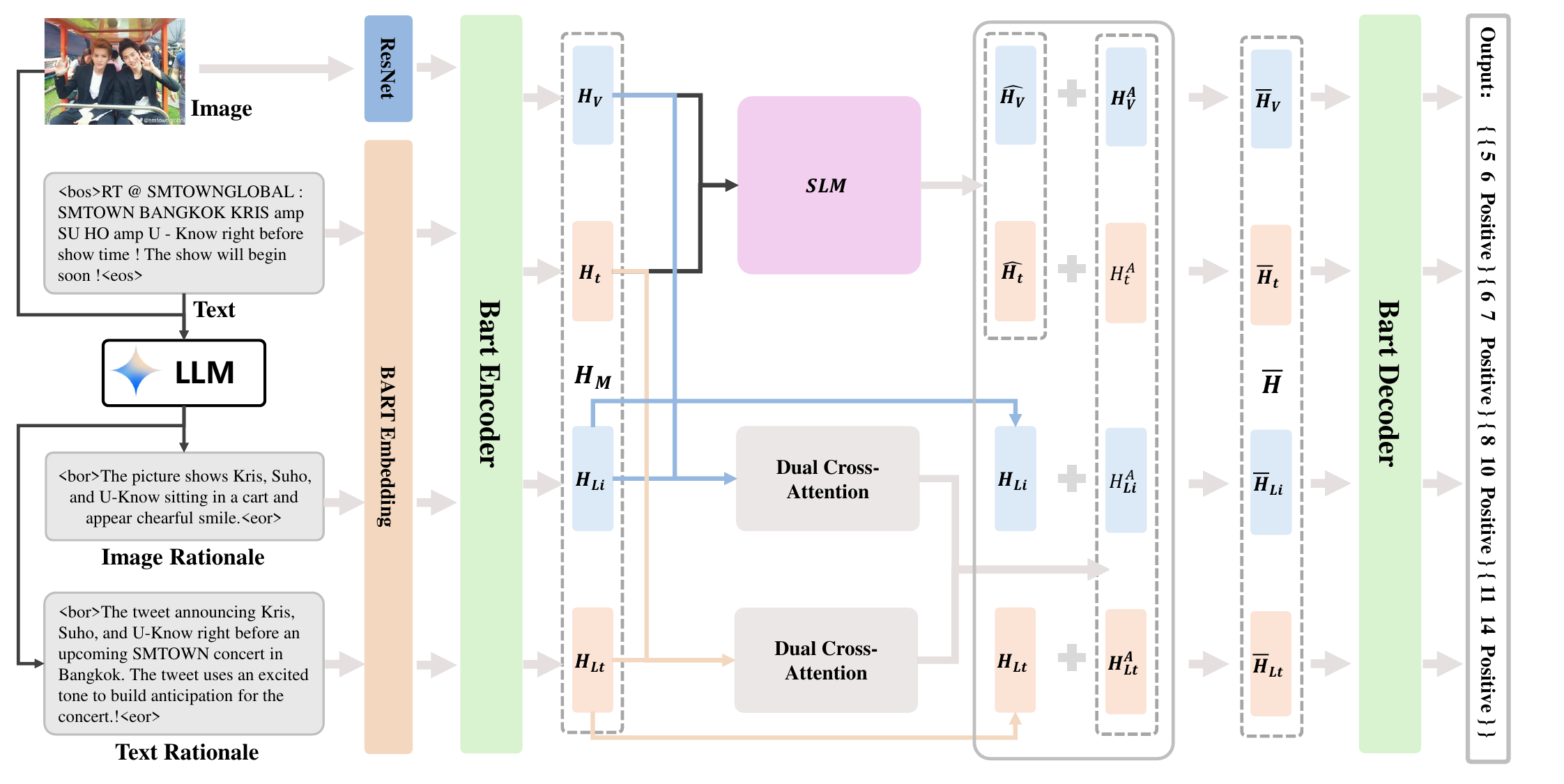}
    \caption{The overview of our proposed aspect-oriented model LRSA.}
    \label{fig2}
\end{figure*}
\section{Proposed Approach}
\subsection{Overview}
\paragraph{Task Definition.}
In formal terms, we examine a image-text pair comprising a sentence \(S = (t1, t2, \cdots, tn)\) with \(n\) words and an image \(V\) . The task aims to extract a sequence (Y) encompassing all aspects and their corresponding sentiment polarities. We define the output as \(Y=[as^s_1,as^e_1,st_1,\cdots,as^s_i,as^e_i,st_i,\cdots,as^s_k,as^e_k,st_k]\), where \(as^s_i\) and \(as^e_i\) denote the begin and end indices for the \(i\)-th aspect, \(st_i\) represents its sentiment polarity, and \(k\) indicates the number of aspects.
\paragraph{Architecture Design.}
As shown in Figure \ref{fig2}, our RASA framework endeavors to incorporate the rationales produced by the LLMs, encompassing both text explanations and image understandings, into the SLMs which is built upon an encoder-decoder framework utilizing BART \cite{lewis2019bart}. To accomplish this objective, we concatenate the image-text pairs, the image rationale and text rationales, which are then fed into the encoder. Additionally, a dual cross-attention mechanism is employed within the SLMs to facilitate fusion. Ultimately, the fused image-text pairs and the rationales of the image and text are once again concatenated and fed into the decoder for result prediction. The following sections elucidate the specifics of our proposed model. 
\paragraph{Data preprocessing.}
 As shown in Figure \ref{fig2}, we input the given images and text into LLMs to obtain the description for the images  \(L_i\) and the explanation for the text \(L_t\). This is then combined with the image rationale and text rationale and inputted into the model as image-text pairs. This will be input into the SLMs as the rationale of images and rationale of texts, with the image-text pairs.
\subsection{ Multimodal Encoder}
We develop a multimodal encoder aimed at capturing comprehensive multimodal representations in this section. We first conduct feature extraction on image-text pairs and image-text rationales. the embeddings of image \(E_V\) are obtained by preprocessing via ResNet \cite{chen2014deepsentibank} or other visual feature extractors. The embeddings of text \(E_T\) and the embeddings of the rationale for images  \(E_{Li}\) and texts \(E_{Lt}\) generated by the LLMs are obtained using the pre-trained BART \cite{lewis2019bart}. We insert <img> and </img> tokens at the beginning and end of the image features, <bos> and <eos> tokens to mark the boundaries of the textual features, and <bor> and <eor> tokens for the rationale features. Following this, we concatenate the multimodal features as \(X=[<img>,E_V,</img>,<bos>,E_{T},<eos>,<bor>,E_{Li},<eor>,<bor>,E_{Lt},<eor>]\), which forms the input for the BART encoder.
In the end, we input X into the BART encoder to derive the multimodal representation. We believe that integrating the image-text pairs and image-text rationales into the same encoder can share parameters and lead to a more efficient and cohesive feature fusion. 
\begin{equation}
    H_M = MBART_E(X), H_M \in \mathbb{R}^{l_m\times d}.
\end{equation}

where \(l_m\) = \(l_i\) + \(l_t\) + \(l_{Li}\) + \(l_{Lt}\),  \(l_i\) represents the number of image slots for the image representation, \(l_t\) denotes the length of the text,  \(l_{Li}\) signifies the length of the image rationale,  \(l_{Lt}\) indicates the length of the text rationale, and \(d\) stands for the hidden dimension.
\subsection{ Dual Cross-Attention Module }
To delve into the nuanced detail of the image-text pair, and to enrich and accurately represent the features, we employ a dual cross-attention mechanism to respectively integrate the image features with image rationale features, and integrate the text features with text rationale features. The detailed explanation is as follows.
First, we divide the multimodal hidden state \(H_M\) obtained from the multimodal encoder into image features \(H_V\), text features \(H_t\), image rationale features \(H_{Li}\), and text rationale features \(H_{Lt}\). We then combine \(H_V\) and \(H_{Li}\) pairwise to form stacked features  \(Z_i\) , combine \(H_V\) and \(H_{Lt}\) pairwise to form stacked features \(Z_t\) , as the input of Dual Cross-Attention Module. Since the formulas required for \(Z_i\) and \(Z_t\) are the same, in the following equations,  \(Z\) represents \(Z_i\) or \(Z_t\),  \(H\) represents \(H_V\) or \(H_{t}\), and \(H_{L}\) represents \(H_{Li}\) or \(H_{Lt}\).
\begin{equation}
    {Z} = 
 \left(
 \begin{matrix}
   {H} \\ {H_{L}} 
  \end{matrix}
  \right) 
  , Z \in \mathbb{R}^{(l_H+l_{H_L})\times d}.
\end{equation}

where \(l_H\) is the length of image feature or text features ,  \(l_{H_{L}}\) is the length of rationale features.

In the Dual Cross-Attention Module. fragment-derived query, key, and value are formulated as follows:
\begin{align}
    &{K}_Z={Z}{W}^K=\binom{{H}\ {W}^K}{{H_L}{W}^K}=\binom{{K}_H}{{K}_{H_L}},\\
&{Q}_Z={Z}{W}^Q=\binom{{H}\ {W}^Q}{{H_L}{W}^Q}=\binom{{Q}_H}{{Q}_{H_L}},\\
&{V}_Z={Z}{W}^V=\binom{{H}\ {W}^V}{{H_L}{W}^V}=\binom{{V}_H}{{V}_{H_L}}.
\end{align}
Then, The Scaled Dot-Product Attention operation is executed as delineated in Eq.(6)
\begin{equation}
    ATTN\!\left(\!{Q}_Z,\!{K}_Z,\!{V}_Z\right)\!=\!softmax\!\left(\!\frac{{Q}_Z\!{K}_Z^T}{\sqrt{d}}\!\right)\!{V}_Z.
\end{equation}
For clarity and simplicity in our derivation, we omit the softmax and scaling functions from the aforementioned equation. This simplification preserves the essence of our attention mechanism. The expanded form is as follows:
\begin{equation}
    \begin{aligned}
        {Q}_Z{K}_Z^T\!\cdot\!{V}_Z\!&=\!\begin{pmatrix}\!{Q}_H\\{Q}_{H_L}\!\end{pmatrix}\begin{pmatrix}\!{K}_H^T&\!{K}_{H_L}^T\!\end{pmatrix}\!\cdot\!\begin{pmatrix}\!{V}_H\\{V}_{H_L}\!\end{pmatrix}
\ \ \;\;\\&=\begin{pmatrix}{Q}_H{K}_H{}^T&{Q}_H{K}_{H_L}{}^T\\{Q}_{H_L}{K}_H{}^T&{Q}_{H_L}{K}_{H_L}{}^T\end{pmatrix}\cdot\begin{pmatrix}{V}_H\\{V}_{H_L}\end{pmatrix}
\qquad\\&=\binom{{Q}_H{K}_H^T{V}_H\!+\!{Q}_H{K}_{H_L}^T{V}_{H_L}}{{Q}_{H_L}{K}_{H_L}^T{V}_{H_L}\!+\!{Q}_{H_L}{K}_H^T{V}_H}\!=\!\begin{pmatrix}\!{H}^A\\{H}^A_{L}\!\end{pmatrix}.
    \end{aligned}
\end{equation}

From the last line of the above equation, it can be concluded that the updated image/text features and image/text rationale features are
\begin{align}
        &{H^A}={Q}_H{K}_H^T{V}_H\!+\!{Q}_H{K}_{H_L}^T{V}_{H_L},
\;\;\,\\&{H^A_L}={Q}_{H_L}{K}_{H_L}^T{V}_{H_L}\!+\!{Q}_{H_L}{K}_H^T{V}_H.
\end{align}

This result illustrates that our dual cross-attention module simultaneously considers the relationship within features and the relationship between image/text features and image/text rationale features. Finally, we get the image features \(H^A_V\), text features\(H^A_t\), image rationale features\(H^A_{Li}\) , and text rationale features \(H^A_{Lt}\) processed by the dual cross-attention module.
\subsection{ Decoder \& Prediction }
After obtaining the fused features in the dual cross-attention module, we add the updated image/text features \(H^A_V\) and \(H^A_t\) to the image/text features \(\hat{H}_V\) and \(\hat{H}_t\) processed by SLMs, and add the updated image/text rationale features \(H^A_{Li}\)  and \(H^A_{Lt}\) to the original image/text rationale features \(H_{Li}\) and \(H_{Lt}\) to get the final features:
\begin{align}
    &\overline{H}_V=H^A_V+\hat{H}_V , \overline{H}_V\in \mathbb{R}^{l_i\times d},\\
&\overline{H}_t=H^A_t+\hat{H}_t \ , \overline{H}_t\in \mathbb{R}^{l_t\times d},\\
&\overline{H}_{Li}=H^A_{Li}+{H}_{Li} , \overline{H}_{Li}\in \mathbb{R}^{l_{Li}\times d},\\
&\overline{H}_{Lt}=H^A_{Lt}+{H}_{Lt} , \overline{H}_{Lt}\in \mathbb{R}^{l_{Lt}\times d}.
\end{align}
To enhance decoder output quality and boost prediction accuracy, we concatenate the final image features  \(\overline{H}_{V}\) and text features  \(\overline{H}_{t}\) with the image rationale features \(\overline{H}_{Li}\) and text rationale features \(\overline{H}_{Lt}\), and the acquired results combined with the preceding decoder output \(Y_{<t}\), serve as input for the BART decoder.
\begin{flalign}
 &\overline{H}=[\overline{H}_{V},\overline{H}_{t},\overline{H}_{Li},\overline{H}_{Lt}]\ , \ \overline{H}\in \mathbb{R}^{l_m\times d},\\
&H^D=MBART_{D}(\overline{H}; Y_{<t}), \\
&\widetilde H = ( \overline{H}_t + E )/2,\\
&P(y)=softmax([\widetilde H;S^d]H^D).
\end{flalign}

\(E\) represents the embeddings of input tokens, while the embeddings of three kind of sentiments and <eos> can be signified as \(S^d\). And the loss function can be expressed thus:
\begin{equation}
    \mathcal{L}_{MABSA}=-\mathbb{E}_{X\sim D}\sum_{t=1}^OlogP(y|Y_{<t},X).
\end{equation}

\section{Experiments}
\subsection{\textbf{\textbf{Experimental Setup} }}
\paragraph{Datasets.}
We selected Twitter2015 and Twitter2017 \cite{yu2019adapting} as our two benchmark datasets. Based on this foundation, we added the description of images as the rationale of image, and the understanding of texts as the rationales of text generated by LLMs to our dataset. However, there are instances where LLMs decline to generate responses for inputs pertaining to politics, violence, and pornography. By utilizing appropriate prompts, we successfully generated principles for the majority of the dataset. Nonetheless, there is a small subset of data for which rationales could not be generated. Table \ref{tab1} outlines the key details of the resulting dataset.
\begin{table}[H]\vspace{-1.0em}\scriptsize
    \centering
    \setlength\tabcolsep{12pt}
     \renewcommand\arraystretch{1.6} 
         \caption{Statistics on two datasets. For A/B, A denotes LLM-processed data quantity, while B indicates the original dataset size.}
    \begin{tabular}{|c|c|c|c|c|} \hline 
         \textbf{Datasets}&  \textbf{Train}&  \textbf{Dev}&  \textbf{Test}& \textbf{Total}\\ \hline 
         \textbf{Twitter2015}&  \textbf{2089}/2101&  \textbf{719}/727&  \textbf{668}/674& \textbf{3476}/3502\\ \hline 
         \textbf{Twitter2017}&  \textbf{1738}/1739&  \textbf{572}/573&  \textbf{585}/586& \textbf{2895}/2898\\ \specialrule{0em}{1pt}{1pt}\hline\specialrule{0em}{1pt}{1pt}
    \end{tabular}
    \label{tab1}\vspace{-1.0em}
\end{table}
\paragraph{Implementation Details.}
Our LLM is Google's latest multimodal large language model "Gemini-1.5 pro" \cite{reid2024gemini}. It can read images and text at the same time and give corresponding answers based on prompts. We input image-text pairs and use specific prompts to generate interpretations of images and understanding of text. We will detail the choice of our prompts in the case study.
We selected two models, VLP \cite{3} and AoM \cite{4}, as our SLMs.  The batch size on Twitter15 datasets is 16, and the batch size on Twitter17 datasets is 18. MABSA task is trained for 35 epochs. Both models share a 7e-5 learning rate and 768 hidden layers.
\paragraph{Metrics.} We use three evaluation metrics to assess the efficacy of various approaches in MASBA and MATE tasks: Micro F1 measure (F1), Precision (P), and Recall (R), while Accuracy (Acc) and F1 evaluate MASC task outcomes.
\subsection{\textbf{Baseline}}
Our model undergoes rigorous evaluation against four distinct methodologies, ensuring a robust comparative analysis.
\paragraph{Models for textual ABSA.}(1) SPAN \cite{hu2019open} introduce a novel span-based extract-then-classify paradigm. (2) D-GCN \cite{chen2020joint} models integrate aspect extraction and sentiment analysis, incorporating syntactic information. (3) BART \cite{yan2021unified}  leverages a pre-trained sequence-to-sequence model  in a cohesive, end-to-end architecture.
\paragraph{Models for MATE.}(1) RAN \cite{wu2020multimoda} RAN [24] aligns text with object regions that show in an image. (2) UMT \cite{yu2020improving} design a Cross-Modal Transformer to guide the final predictions. (3) OSCGA \cite{wu2020multimodal} concentrates on the interconnected relationships between visual elements and their corresponding entities.
\paragraph{Models for MASC.}(1) ESAFN \cite{8926404} study entity-level multimodal sentiment classification. (2) TomBERT \cite{yu2019adapting} apropose a multimodal BERT architecture to model intra-modality dynamically. (3) CapTrBERT \cite{Khan} expands the textual input capacity for the language processing system.
\paragraph{Models for MABSA.} (1) UMT-collapse \cite{yu2020improving}, OSCGA-collapse \cite{wu2020multimodal}  and RpBERT-collapse \cite{sun2021rpbert} approaches the JMASA task by utilizing a collapsed label framework.  2) UMT+TomBERT, OSCGA+TomBERT integrate MATE and MASC methodologies in a sequential framework. 3) JML \cite{2} first implemented MABSA in a unified framework. 4) CMMT \cite{yang2022cross} introduced a cross-modal multi-task transformer (CMMT) for MABSA. 5) VLP-MABSA \cite{3} built a generative multimodal architecture for downstream MABSA tasks. 6)AoM \cite{4}  prioritizes the extraction of aspect-relevant textual and visual elements. 
Notably, we reproduced the model using the authors' specified parameters, but the experimental results did not match the outcomes reported in the paper.

\begin{table*}[h]\scriptsize%
\centering
\setlength\tabcolsep{6pt} 
\caption{\label{Tab2}
MABSA task results from both Twitter datasets.  *indicates that we replicated the experiments locally using the code and parameters provided by the paper's authors and selected the optimal results. We use bold font for the best results and underlined font for the second-best results.
}
\begin{tabular}{lllllllll} 
\hline
\specialrule{0em}{1pt}{1pt}
 &\textbf{Methods}& \multicolumn{3}{c}{\textbf{Twitter2015}} & & \multicolumn{3}{c}{\textbf{Twitter2017}}\\ \specialrule{0em}{1pt}{0pt} \cline{3-5} \cline{7-9}
 \specialrule{0em}{1pt}{1pt}
 & &  P& R& F1 & & P& R&F1\\ 
\hline
\specialrule{0em}{1pt}{1pt}
 &SPAN \cite{hu2019open} & 53.7& 53.9& 53.8 & & 59.6& 61.7&60.6\\ 

 Text-based&D-GCN \cite{chen2020joint} & 58.3& 58.8& 59.4 & & 64.2& 64.1&64.1\\
 &BART \cite{yan2021unified} & 62.9& 65.0& 63.9 & & 65.2& 65.6&65.4\\ 
 \specialrule{0em}{1pt}{1pt}
 \hline
\specialrule{0em}{1pt}{1pt}
 &UMT+TomBERT \cite{yu2020improving,yu2019adapting} & 58.4& 61.3& 59.8 & & 62.3& 62.4&62.4\\ 
 &OSCGA+TomBERT \cite{wu2020multimodal,yu2019adapting} & 61.7& 63.4& 62.5 & & 63.4& 64.0&63.7\\ 
 &OSCGA-collapse \cite{wu2020multimodal} &  63.1& 63.7& 63.2 & & 63.5& 63.5&63.5\\ 

 & RpBERT-collapse \cite{sun2021rpbert} & 49.3& 46.9& 48.0 & & 57.0& 55.4&56.2\\ 
 Multimodal& UMT-collapse & 61.0& 60.4& 61.6 & & 60.8& 60.0&61.7\\ 
 & JML \cite{2} & 65.0& 63.2& 64.1 & & 66.5& 65.5&66.0\\ 
 & VLP-MABSA* \cite{3} & 65.1& 68.3& \underline{66.6} & & 66.9& 69.2&68.0\\ 
 & CMMT \cite{yang2022cross} & 64.6& 68.7& 66.5 & & 67.6& \underline{69.4}&\underline{68.5}\\ 
 & AoM* \cite{4} & 65.3& 67.3& 66.3 & & 66.5& 67.3&66.9\\
 & LRSA on VLP (Ours) & \underline{66.7}& \textbf{69.7}& \textbf{68.2} & & \textbf{68.2}& \textbf{70.2}&\textbf{69.2}\\ 
 & LRSA on AoM(Ours) & \textbf{67.2}& \underline{69.2}& \textbf{68.2} & & \underline{67.8}& 68.8&68.3\\ 
  \specialrule{0em}{1pt}{1pt}
 \hline
   \specialrule{0em}{1pt}{1pt}\vspace{-1.0em}
\end{tabular}

\end{table*}

\subsection{\textbf{Main Results}}

\paragraph{Results of MABSA Task:}
Table \ref{Tab2} displays the outcomes derived from the MABSA task. Firstly, our LRSA model significantly surpasses all text-based models, indicating the substantial contribution of incorporating image modality information and conducting text analysis of image modality in our model. Secondly, we use two small models, VLP and AoM, as the baseline of the model for experiments.  The comparison revealed that LRSA exhibited a 1.6\% and 1.2\% improvement in F1 metric compared to VLP,  LRSA exhibited a 1.9\% and 1.4\% improvement in F1 metric compared to AoM, demonstrating its generalizability and applicability to most pre-trained models for MABSA. Lastly, LRSA  outperforms all other multimodal methods across all evaluation metrics, highlighting the effectiveness of augmenting the small model with additional information about image-text pairs generated by the large model in boosting the overall model performance.

\paragraph{Results of MATE Task:}
As shown in Table \ref{tab3}, LRSA on two SLMs outperforms the majority of contemporary models, The performance of LRSA on VLP is 0.5\% and 1.1\% higher than VLP.This demonstrates that the additional information from image-text pairs generated by LLMs can effectively enhance the ability of SLMs to recognize aspect terms. 
\begin{table}[H]\vspace{-1.0em}\scriptsize 
    \centering
    \setlength\tabcolsep{3pt} 
    \caption{Performance comparison of MATE and MASC methodologies. }
    \begin{tabular}{ccccccc|ccccc}
    \hline
\specialrule{0em}{1pt}{1pt}
         \multicolumn{7}{c|}{\textbf{MATE}}&  \multicolumn{5}{c}{\textbf{MASC}}\\
         \hline
\specialrule{0em}{1pt}{1pt}
         \textbf{Method}&  \multicolumn{3}{c}{\textbf{Twitter2015}}&  \multicolumn{3}{c|}{\textbf{Twitter2017}}&  \textbf{Method}&  \multicolumn{2}{c}{\textbf{Twitter2015}}& \multicolumn{2}{c}{\textbf{Twitter2017}} \\\specialrule{0em}{1pt}{0pt} \cline{2-4} \cline{5-7} \cline{9-10} \cline{11-12}
 \specialrule{0em}{1pt}{1pt}
         &  P&  R&  F1&  P&  R&  F1&  &  ACC& F1& ACC&F1\\
         \hline
\specialrule{0em}{1pt}{1pt}
         RAN&  80.5&  81.5&  80.0&  90.7&  90.7&  90.0&  ESAFN&  73.4& 67.4& 67.8&64.2\\
         UMT&  77.8&  81.7&  79.7&  86.7&  86.8&  86.7&  TomBERT&  77.2& 71.8& 70.5&68.0\\
         OSCGA&  81.7&  82.1&  81.9&  90.2&  90.7&  90.4&  CapTrBERT&  78.0& 73.2& 72.3&70.2\\
         JML&  83.6&  81.2&  82.4&  \textbf{92.0}&  90.7&  91.4&  JML
&  78.7& -& 72.7&-\\
         VLP&  83.6&  \textbf{87.9}&  85.7&  90.8&  \underline{92.6}&  91.7&  VLP
&  78.6& 73.8& 73.8&71.8\\
 AOM& 83.4& 85.9& 84.6& 89.7& 90.8& 90.2& AOM
& 78.3& 73.5& \underline{74.2}&72.2\\
 Ours(VLP)&\textbf{85.4} &87.0 &\textbf{86.2} &\underline{91.3} &\textbf{94.0} &\textbf{92.6} & Ours(VLP)
&\textbf{79.6} &\underline{74.4} &\textbf{74.7} &\textbf{73.3}\\
 Ours(AoM)& \underline{84.1}& \underline{87.7}& \underline{85.8}& \textbf{92.0}& \underline{92.6}& \underline{92.3}& Ours(AoM)& \underline{79.5}& \textbf{75.9}& \underline{74.2}&\underline{73.2}\\
 \hline
\specialrule{0em}{1pt}{1pt}
    \end{tabular}

    \label{tab3}
\end{table}
\paragraph{Results of MASC Task:}
The MASC performance outcomes is presented in Table \ref{tab3}. LRSA on VLP outperforms the VLP results by 1\% in accuracy on Twitter2015,
LRSA on VLP outperforms the AoM results by 0.5\% in accuracy on Twitter2017, LRSA on AoM outperforms the VLP results by 2.1\% in F1 score on Twitter2015, LRSA on VLP outperforms the AoM results by 1.1\% in F1 score on Twitter2017. This shows that LRSA helps the SLMs better analyze the sentiment of the aspect via supplementary rationale data furnished by the LLMs..

\subsection{Ablation Study}
We perform ablation studies on LRSA to assess module effectiveness. Table \ref{tab4} illustrates results, confirming the full LRSA model's consistent superiority across all tasks.

\paragraph{W/O dual cross-attention} We experimented with unilateral cross-attention, in which only the rationale features were integrated into the image/text pair features while leaving the rationale features unaltered. The findings indicated a slight reduction in performance. We posit that as the rationale features are concatenated with the image/text pair features as decoder input during the decoding stage, it is crucial to employ the attention mechanism to eliminate noise and accentuate essential features within the rationale features.

\paragraph{W/O cross-attention} performs worse after removing dual cross-attention module. The necessity of employing the dual cross-attention mechanism to integrate the features of image-text pairs and rationale is substantiated. This mechanism effectively filters out noise from the features of image-text pairs and rationale, amplifies essential features, and facilitates the extraction of intricate information.
\paragraph{W/O Concentration} The decoder stage receives input only from the image/text features fused with rationale features, without concatenating the rationale features. The resulting performance is lower, indicating that in the generation stage, providing additional rationale features information assists the SLMs in making more accurate judgments and generating improved results.
\begin{table}[H] \vspace{-1.0em} \scriptsize
    \centering
    
        \setlength\tabcolsep{4pt} 
    \caption{Results of Ablation Study}
    \begin{tabular}{c|cccccc|cccccc}
        \hline
\specialrule{0em}{1pt}{1pt}
         &  \multicolumn{6}{c|}{\textbf{Ours on VLP}}&  \multicolumn{6}{c}{\textbf{Ours on AoM}}\\
            \specialrule{0em}{1pt}{0pt} \cline{2-4} \cline{5-7} \cline{8-10} \cline{11-13}\specialrule{0em}{1pt}{1pt} 
         \textbf{Methods}&  \multicolumn{3}{c}{\textbf{Twitter2015}}&  \multicolumn{3}{c|}{\textbf{Twitter2017}}&  \multicolumn{3}{c}{\textbf{Twitter2015}}& \multicolumn{3}{c}{\textbf{Twitter2017}}\\\specialrule{0em}{1pt}{0pt} \cline{2-4} \cline{5-7} \cline{8-10} \cline{11-13}\specialrule{0em}{1pt}{0pt} 
         &  P&  R&  F1&  P&  R&  F1&  P&  R&  F1& P& R&F1\\
             \hline
\specialrule{0em}{1pt}{1pt}
         Full&  \textbf{66.7}&  \textbf{69.7}&  \textbf{68.2}& {68.2} & \textbf{70.2} & \textbf{69.2} & \textbf{67.2} & \textbf{69.2} & \textbf{68.2} & \textbf{67.8}& \textbf{68.8}&\textbf{68.3}\\
         w/o D-C-A& {66.4} &68.7  & 67.5 & 68.0 & 69.8 & 68.9 & 66.9 &68.1  &67.5  &66.8 &\textbf{68.8} & 67.8\\
         w/o C-A&66.3  &67.1  &66.7  &\textbf{68.4}  &68.2  &68.3  &\textbf{67.2}  &67.7  &67.4  &67.0 &67.3 &67.0\\
         w/o  concat&65.9  &67.6  &66.7  &67.3  &69.5  &68.4  &{65.8}  &68.8  &67.3  &66.7 &68.5 &67.6\\
                      \hline
\specialrule{0em}{1pt}{1pt}
    \end{tabular}

    \label{tab4}\vspace{-1.0em}
\end{table}

\subsection{Discussion}
Our model incorporates additional rationale information generated by LLMs and inputs that information into SLMs to enhance judgment. Thus, the quality of the provided rationale information is of utmost importance. Selecting suitable prompts to guide the LLMs in generating corresponding rationales poses a significant question. To address this, we conducted experiments from two perspectives: the sentiment analysis degree within the LLMs and the length of the generated rationales. A detailed description of these experiments follows:
\begin{table}[H]\scriptsize
    \vspace{-1.0em}
    \centering
        \caption{ three types of prompts for rationale generation}
    \begin{tabular}{|p{2cm}|p{10cm}|} \hline 
    
    \specialrule{0em}{1pt}{1pt}
         \textbf{Cases}& \textbf{prompt}\\ \hline 
         \specialrule{0em}{1pt}{1pt}
         prompt with- & Q1:Explain the text above.\\
         
         out any task-&Q2:Explain the picture.\\
          related hints&
         The Answer should be no more than 140 words.\\ \hline 
         \specialrule{0em}{1pt}{1pt}
         prompt with& you will perform a aspect-based sentiment analysis task with me, You \\
          task-related&are an assistant for the task.\\
           hints&For the above pictures and text, Q1: explain the above text, Q2: explain the above picture and expressions if there are character in the picture.\\
           &The answer should be no more than 140 words.\\ \hline 
         \specialrule{0em}{1pt}{1pt}
         prompt with 
& First, explain the aspect of Aspect-based sentiment analysis. 
Then you \\
detailed task-&will perform an aspect-based sentiment analysis task with me, \\
related hints &You are an assistant for the task. 
This is the example of the task: \\
and allow &Tweet: "On the scene of a robbery at Regions Bank at 4003 University\\ LLMs to give&Drive. Officers have K - 9 looking for any traces." \\
opinions. &In this example, the aspects are "4003 University Drive" and "K - 9", the sentiment of them is neutral. 
Don't analyze the example. \\
&Then for the task question, given the following tweet, analyze this sentence from the sentiment analysis perspective to better help me find aspect words and determine their sentiment. \\
&Tweet: “RT @ Gabriele Corno: Sunset over Albemarle Sound in Kill Devil Hills, North Carolina by Scott Evers.”\\
&Please note aspects in tweets may be only one, or multiple, not all nouns are aspects, but aspects must consist of one or more nouns in the tweet as a subject. Choose One or Two aspects then explain. There must be an explanation and should be no more than 140 words. Don't give ambiguous opinions.\\ \hline
\specialrule{0em}{1pt}{1pt}
    \end{tabular}
    \label{tab5}
    \vspace{-2.0em}
\end{table}

\paragraph{the sentiment analysis degree within the LLMs }
In our model, LLMs are responsible for generating rationale information to aid the judgment of the SLMs. To this end, we have evaluated three types of prompts shown in Table \ref{tab5}. The first type of prompt does not offer any task-related hints; instead, it simply requests an explanation of the image-text pairs. The second type of prompt notifies the LLMs that we are conducting sentiment classification tasks, and it prompts them to provide explanations for both the text and image, along with an analysis of the sentiment conveyed in the image. The third type of prompt informs the LLMs that we are carrying out aspect-based sentiment analysis tasks and provides illustrative examples to clarify the requirements of the task. Subsequently, we request the LLMs to analyze a sentence from a sentiment analysis perspective to assist in identifying aspect words and determining their associated sentiment. Our experiments were conducted using the Twitter2015 dataset and considered the same length of rationale as the experimental background. The result shown in Table \ref{tab6}, indicate that the second prompt yield the most favorable outcomes. We attribute the subpar performance of the first prompt to the excessively broad rationales generated, which failed to specifically address the task of MABSA and introduced significant irrelevant noise. Nevertheless, this prompt still provided supplementary information for the SLMs. Conversely, the rationales generated by the third prompt were overly specific, leading the LLMs to provide inaccurate opinions regarding aspects and sentiments. Consequently, this misled the SLMs and resulted in erroneous judgments. These observations partially validate our approach, which posits that the LLMs are not explicitly responsible for determining aspects and sentiments. Instead, they are solely employed for extracting fine-grained information from the image-text pairs to be used by the SLMs in judgment.
\begin{table}[H]\scriptsize 
    \centering
        \setlength\tabcolsep{4pt} 
     \renewcommand\arraystretch{1.2} 
    \caption{Result of experiments from two perspectives: the sentiment analysis degree within the LLMs and the length of the generated rationales.}
    \begin{tabular}{p{9cm}|ccc}
                 \hline
\specialrule{0em}{1pt}{1pt}
          &  \multicolumn{3}{c}{\textbf{Twitter2015}}\\\cline{2-4}
         \textbf{Cases} &  P&  R& F1 \\
                       \hline
\specialrule{0em}{1pt}{1pt}
         prompt without any task-related hints&65.5  &67.0  &66.3 \\
                        \hline
\specialrule{0em}{1pt}{1pt}
         prompt with task-related hints&\textbf{66.7}  &\textbf{69.7}  &\textbf{68.2} \\
                        \hline
\specialrule{0em}{1pt}{1pt}
         prompt with detailed task-related hints  and allow LLMs to give their own insights&63.4  &64.4  &63.9 \\
                        \hline
\specialrule{0em}{1pt}{1pt}
  The rationale's length corresponds to that of the text-image pair &66.2 &68.5 &67.3\\
                 \hline
\specialrule{0em}{1pt}{1pt}
  The rationale's length doubles that of the text-image pair &\textbf{66.7 } &\textbf{69.7}  &\textbf{68.2}\\
                 \hline
\specialrule{0em}{1pt}{1pt}
  The rationale's length quadruples the text-image pair's length &65.0 &65.9 &65.5\\
               \hline
\specialrule{0em}{1pt}{1pt}
    \end{tabular}

    \label{tab6}\vspace{-1.0em}
\end{table}
\paragraph{the length of the rationales}
The extent of rationales input into the SLMs is also a significant factor affecting the model's efficacy. We explored three different lengths: equal to the average length of the text-image pair feature, twice to the average length of the text-image pair feature, and four times the average length of the text-image pair. When the generated rationale's length falls below the text-image pair's average extent, it will often only generate a sentence that repeats the text content. For our experiments, we utilized the Twitter2015 dataset and considered the second prompt as the experimental background. The results, shown in Table \ref{tab6},  indicate that the best effect is achieved when the total length of the text-image rationale feature is twice the average length of the text-image pair feature. We posit that when the text-image feature is equal to length of the text-image pair feature, it provides insufficient information to the SLMs, leading to an unclear effect. Similarly, when the text-image feature is four times as long as the text-image pair feature, the capability of the SLMs is limited, making it challenging to accurately select and process excessive information. Additionally, an abundance of principle information may contain irrelevant noise, thus affecting the selection of the SLMs.

\section{Conclusion}
In this Paper, we introduces an innovative LRSA that combines the decision-making capabilities of SLMs with the additional information provided by LLMs for MABSA, by using rationales generated by LLMs for SLMs to reference. And to better capture fine-grained information and enhance feature representation, we employed a dual cross-attention mechanism to separately fuse image features with the rationales features of LLMs' understanding of images, and text features with the rationales features of LLMs' understanding of texts. Experimental results on two benchmark datasets show our approach surpasses prior methods.
\section*{Acknowledgements}
This work was partially supported by JSPS KAKENHI Grant Number JP23K28092 and the National Key Technology Research and Development Program of China under grant No. 2022YFB3105401.

%
%
%

\bibliographystyle{splncs04}

\bibliography{mybibliography}

\end{document}